# Application of Kullback-Leibler Metric to Speech Recognition

## Igor Bocharov, Pavel Lukin

**Nizhny Novgorod Linguistic University, Russia**

Article discusses the application of Kullback-Leibler divergence to the recognition of speech signals and suggests three algorithms implementing this divergence criterion: correlation algorithm, spectral algorithm and filter algorithm. Discussion covers an approach to the problem of speech variability and is illustrated with the results of experimental modeling of speech signals. The article gives a number of recommendations on the choice of appropriate model parameters and provides a comparison to some other methods of speech recognition.

## Introduction

The continuously growing role of computers and their widespread occurrence in everyday life has brought up a grave need for an intuitive user-friendly interface with various automatic systems. As soon as speech has always been one of the fastest and most natural means of human communication, it is reasonable that the development of automatic speech recognition systems became one of the most topical tasks of today's computer science. The history of speech analysis and recognition is quite lengthy by now [1], however, no widely accepted tool with built-in speech input has yet been introduced. The only success in this respect is probably the voice-activated control in mobile systems. Any wider implementation of speech interface is tied up by the lack of robust and effective recognition algorithms with moderate computation requirements.

The recognition approach suggested in the article is based on the least information divergence criterion (LID) [2] applied to information characteristics of the input signals, which allows to point to information-theoretic optimality of algorithm implementations. Feature extraction uses effective and well-studied methods. The practical implementation offers three algorithms: a frequency domain spectral algorithm, and two time domain ones: correlation algorithm and filter algorithm. In order to level the variable duration of input signals a computationally-effective scheme of signal alignment was introduced. Discussion coves the design and experimental investigation of LID algorithms, parameter optimization and a comparative effectiveness analysis of algorithms one with another and with other existing methods.

## Least information divergence in signal recognition

The problem of recognition of $R$ Gaussian signals was formulated and solved in [3] under the assumption of Kullback-Leibler metric $I_n$ [4] as the measure of mutual entropy between two distributions $f_x$ and $f_r$:

$$\mathrm{I}_n\left[f_x \mid f_r\right] \stackrel{\Delta}{=} \iint_n \ln\left(\frac{f_x(\vec{x})}{f_r(\vec{x})}\right) f_x(\vec{x}) d\vec{x}. \tag{1}$$

Information divergence between two stochastic processes $x$ and $r$ of length $n$ defined by KL metric (1) can be expressed in the closed form by the autocorrelation matrices of $x$ and $r$:

$$I_n(x \mid r) = \frac{1}{2} tr\left(\mathbf{K}_r^{-1} \mathbf{K}_x\right) + \frac{1}{2} \ln\left(\frac{\det \mathbf{K}_r}{\det \mathbf{K}_x}\right) - \frac{n}{2} \tag{2}$$

where $tr$ is matrix trace, and $det$ is matrix determinant.

In practice, the computation of (2) for real signals faces the challenge of a priori uncertainty of their correlation matrices. Final choice of matrix estimation generally offers three different approaches for mutual entropy computation:

1. Direct computation from correlation matrix estimates.
2. Frequency domain computation based on the power spectral density of the signals.
3. Whitening filtration (WF) scheme based on autoregressive model of speech signal.

Regardless of the approach, the design of automatic recognition system is assumed to have classical architecture [5], that is, actual recognition is preceded by the preparation of a feature database for each recognized word. At recognition step the input word is processed analogously to the previously built database in order to extract the features, which are then compared to the features from the database. The result of this multi-channel computation is the decision in favor of one of the database words, which has the least information divergence from the input signal. Each particular approach to the computation of the metric defines its own set of features and the type of decision rule for the described architecture. Let us consider each of them in detail.

### Correlation algorithm

The correlation interpretation of the method is based on estimation of autocorrelation matrices of the analyzed signals. The major advantage of this approach is its direct derivation from LID and consequent absence of the limit process in the formulation of decision statistic. Besides it is applicable to nonstationary processes, in particular, to speech signals. Decision statistic in case of correlation-based LID is the following:

$$g_{x,r} = tr\left(\hat{\mathbf{K}}_r^{-1}\hat{\mathbf{K}}_x\right) - \ln\left(\det \hat{\mathbf{K}}_x\right) \rightarrow \min_r \bigg|_{r=\overline{1,R}} \quad (3)$$

where: $\hat{\mathbf{K}}_r^{-1}$ is the inverse autocorrelation matrix estimate for a signal from the dictionary, $\hat{\mathbf{K}}_x$ is the autocorrelation matrix estimate for the input word, $R$ is dictionary size.

Therefore, autocorrelation matrix estimates of compared signals become features for future recognition. Matrix estimates are formed in a running window, e. g. for signal $x$ of length $n$ the autocorrelation matrix estimate of order $P$ can be found as follows:

$$\hat{\mathbf{K}} = \frac{1}{w}\sum_{i=1}^{w}\vec{x}_P \cdot \vec{x}_P^T \quad (4)$$

where $\vec{x}_P$ is a column-vector obtained by extracting $P$ samples from $x$ starting from $(i \cdot P)$th sample, $w$ is the number of windows, equal to $n/P$;

The reference feature dictionary consists of inverse autocorrelation matrices of all $R$ signals. Note, that computationally intensive matrix inversion is required only to build up the database and is not used in recognition, which involves only estimation of $\hat{\mathbf{K}}$ in (4) and the computation of decision statistic (3) for all $R$ autocorrelation matrices in the dictionary. Final decision is in favor of such word in the dictionary that has the least value of (3). Parameter subject to possible optimization is the order $P$ of matrix $\mathbf{K}$.

The field of application of autocorrelation methods is bound by the computational complexity of linear algebra operations. In particular, statistic (3) requires an inversion of a matrix and a matrix determinant. However, these two operations can be combined into single inversion algorithm based on Crout's LU-decomposition [6] and implemented as a linear algebra SIMD microprocessor instruction of Intel's Pentium IV [7]. This algorithm allows to find the inverse of a matrix in parallel with its determinant. An original and effective algorithm of estimating the inverse autocorrelation matrix for autoregressive signals exploits the idea of a triple matrix product and was suggested in [8]. It is worth mentioning that computational complexity of the algorithm can be decreased further by implementing fast algorithms for special matrix types (Toeplitz matrices) for example Levinson-Durbin algorithm [9].

### Spectral algorithm

The development of Fast Fourier Transform (FFT) sharply reduced the computational requirements of traditional algorithms and enabled a faster yet effective discerning of signals in the frequency domain. The correlogram estimate of signal's power spectral density allows to express the information divergence between two signals via their spectral characteristics. At the limit, when $n \to \infty$ the autocorrelation matrix reflects the information of signal's spectrum, which allows to formulate an asymptotically optimal decision statistic for two signals in the frequency domain:

$$g_{x,r} = \frac{1}{F} \sum_{f=1}^{F} \left( \frac{\hat{G}_x(f)}{\hat{G}_r(f)} + \ln \frac{\hat{G}_r(f)}{\hat{G}_x(f)} \right) \to \min_r \bigg|_{r=\overline{1,R}} \tag{5}$$

where $\hat{G}_x(f)$ is the sample estimate of power spectral density of the recognized signal $x$; $\hat{G}_r(f)$ is the sample estimate of power spectral density of signal $r$ from the dictionary; $F$ is half the sampling rate; $R$ is dictionary size.

Feature to be extracted from the signal is therefore the estimate of its power spectral density (PSD). The estimate is made by FFT in a running Hamming window; parameter subject to possible optimization is FFT window's length. In order to maintain computational efficiency experimental computations were based on classical FFT [9] with window size equal to a power of two. A possible evolution of this method could be better spectral estimates for (5) that have better account for the structural differences in speech signals.

The feature database consists of power spectrum estimates for all $R$ signals in the dictionary. Recognition step involves computation of input signal's power spectrum estimate and the decision statistic (5) for the signals in the dictionary. Decision is in favor of such word from the dictionary that has the least value of (5).

### Filter algorithm

Initial premise underlying the idea of whitening filtration (WF) is the autoregressive (AR) model of speech signal. This approximation became firmly established in speech synthesis as a fair description of the vocal tract behavior under the assumptions of lossless tube model [10]. The idea of digital filtration based on filters with simple transversal topology allows to implement a recognizer running on cheap high-performance DSP [11].

Decision statistic for the signals obtained by means of linear filtration of "white" Gaussian noise is formulated on the assumption of AR nature of analyzed signals (normalized to the variance of the generating process):

$$g_{x,r} = \frac{s_{x,r}^2}{s_r^2} + \ln \frac{s_r^2}{s_{x,r}^2} \to \min_r \bigg|_{r=\overline{1,R}} \tag{6}$$

where: $s_{x,r}^2$ is the dispersion of uncompensated remainder for signal $x$ at the output of WF tuned to $r$ th signal in the dictionary; $s_r^2$ is the variance of the excitation white Gaussian noise in AR model of signal $r$; $R$ is the dictionary size.

The value in (6) is to a constant in accordance with the value of information divergence between signal $x$ and signal $r$ from the dictionary in Kullback-Leibler metric (1), and therefore conforms to LID criterion.

The algorithm is implemented as a set of parallel transversal whitening filters $WF_{1..R}$ of order $P$ each that are inverse to the respective forming filters and are formed from signal's AR coefficients. The feature extracted from the signal is therefore a $P$-vector of AR coefficients, which is calculated by minimizing linear prediction error with respect to its variance. In this regard a method of special interest is Burg's algorithm [9] that allows to construct filters with high dynamical characteristics; the choice of filter order $P$ is an optimization problem of its own. The database consists of $R$ AR vectors that are used in recognition to construct respective whitening filters. Decision is based on the minimum of decision statistic (6) in $R$ recognition channels.

## Experimental setup

The implementation of automatic system for the task of separate word recognition falls into two time-independent parts: dictionary creation (training) and actual recognition. The structural diagram of the system is shown on figure 1.

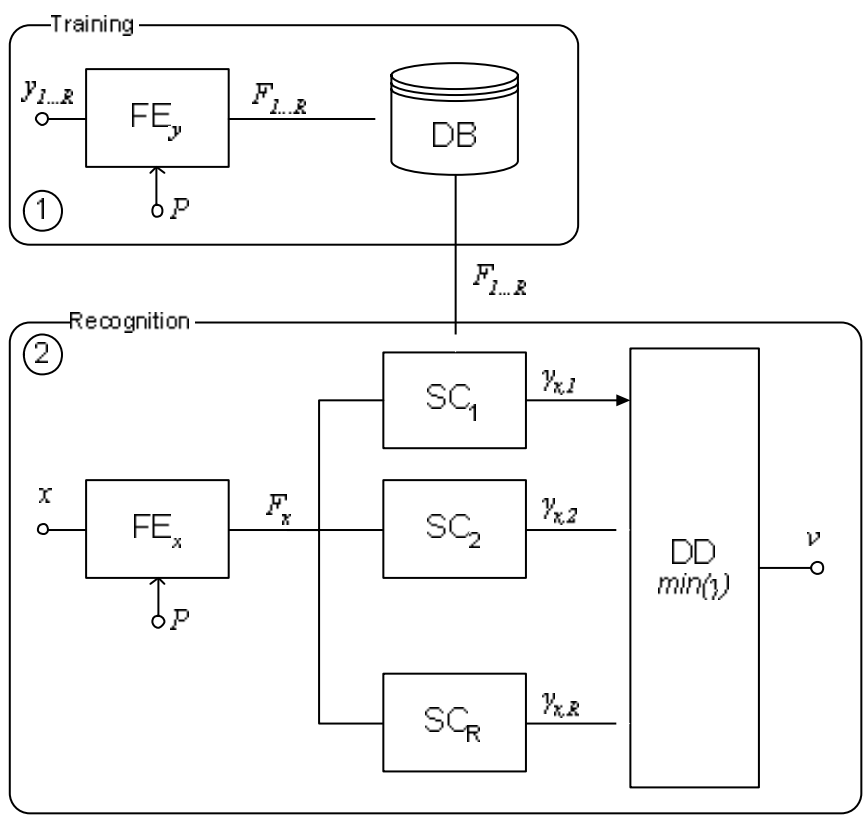

input signal is subject to feature extraction by the same algorithm as in the database step. Note, that in general, the value of $P$ in $\text{FE}_y$ used in dictionary creation can be different from the one in $\text{FE}_x$ used in recognition. Final decision is obtained in Decision Device $\text{DD}$, where the winning word is the one with the least information divergence $\boldsymbol{g}_{x,r}$ from the input signal at the outputs of Statistic Calculators $\text{SC}_{1..R}$.

### Speech variability

The increasing complexity of speech recognition algorithms is strongly related to speech variability i.e. to significant parametric differences in the pronunciations of one word. An approach that assumes each time-limited non-stationary signal as an observation of an infinitely long stationary signal is seriously constrained in the efficiency of its practical implementations, because in practice speech production is only piecewise time-stationary. Theory and practice of speech signal analysis [12] indicate so far that human speech tract remains stationary approximately for 20 ms.

What are the ways to allow for the non-stationary (phonemic) signal behavior? In case of small dictionary sizes with a priori known structure feature extraction can be applied to whole words, without preliminary segmentation. A theoretic background of this approach is the argument that analyzed signal (generally nonstationary within the observation interval) can be treated as an instantaneous sample from an infinitely long stationary signal. Thus, the features extracted from the analyzed signal represent the integral characteristics of this stationary signal. In the frequency domain this is equivalent to sequential windowing of the analyzed signal and accumulation of its formants, where the resulting characteristic integrates all analyzed signal formants. The input of each formant is proportional to its length, i.e. each input is automatically normalized with respect to overall input to the integral characteristic. This argument is based on tight interconnection between spectral, correlation and autoregressive characteristics and remains true for correlation and autoregressive features extracted from the signal as a whole. The described approach to the problem of speech variability is quite sufficient to find out the characteristics of recognition metric.

## Experimental results

The subject of experimental investigation was the performance of speech recognition system with a limited dictionary (10-15 words) implementing the described LID methods. In most cases such systems are parts of automatic control systems that provide voice-interface to other complex devices. The dictionary of the experimental system was built using Russian words (ten digits from zero to nine). Sampling frequency (8kHz) and quantization (8bit) were chosen in view of the most typical values of these parameters in the existing speech recognition tools. The standard channel of built-in PC sound unit and the low-price microphone seem to have made the experiment more life-realistic. As a result the signals were masked with an additive non-stationary noise from PC units at about 18dB and an additive noise background at about 16dB. Reference dictionary was created by one speaker and included 10 digits (from "0" to "9", one sample of each). Experimental database was created by the same speaker and included a 100 of equally intoned samples for each digit from "0" to "9". All signals had allophone positional variability and length variability not exceeding 120%.

The purpose of investigation was estimation of probability (event frequency) $w$ of correct word recognition depending on the model parameter value; the probability was computed after a sequence of a 100 trials as follows:

$$w = \frac{K_{corr}}{K_{tot}}$$

where $K_{corr}$ is the amount of correctly recognized words in the whole sequence; $K_{tot}$ is the total amount of trials.

The results of experiments are shown on figures below.

Figure 2 shows how the probability of recognition of digit "4" (correlation algorithm) relates to the order of the matrix.

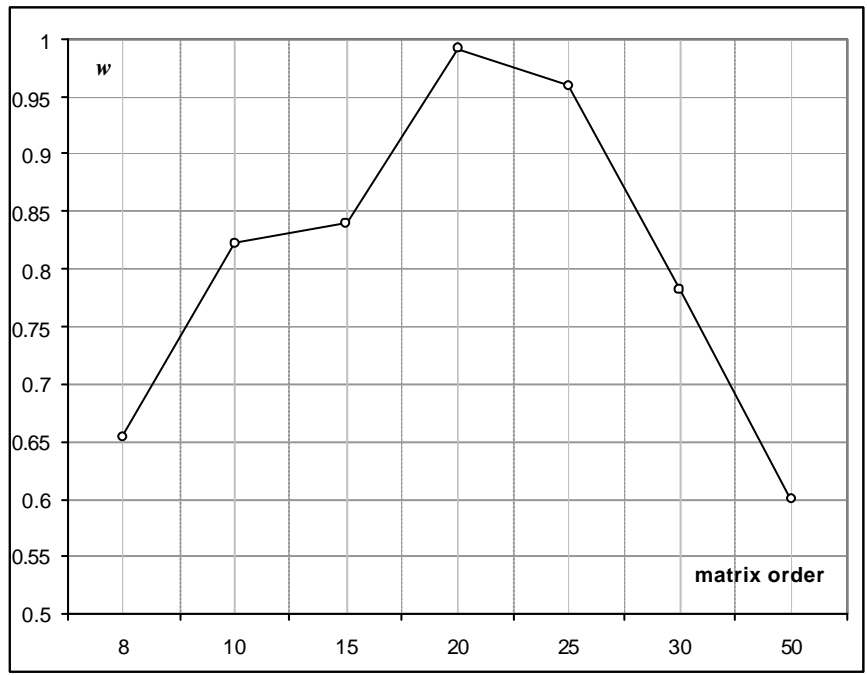

**Figure 2**  **Probability of correct recognition (correlation algorithm)**

It can be seen from the figure that the probability of correct recognition has a distinct maximum at 20-25, which represents an optimal value for the size of correlation matrix for the signal in question. The very slight decrease in the value of *w* as matrix order is increased is a good characteristic for the reliability of most matrix orders. As one would expect, correlation algorithm has shown the best recognition performance, in full consistency with its theoretic background. However, the price paid for this are higher requirements to computational resources (both CPU and memory).

Figure 3 shows how the probability of recognition of digit "4" (spectral algorithm) relates to FFT window length.

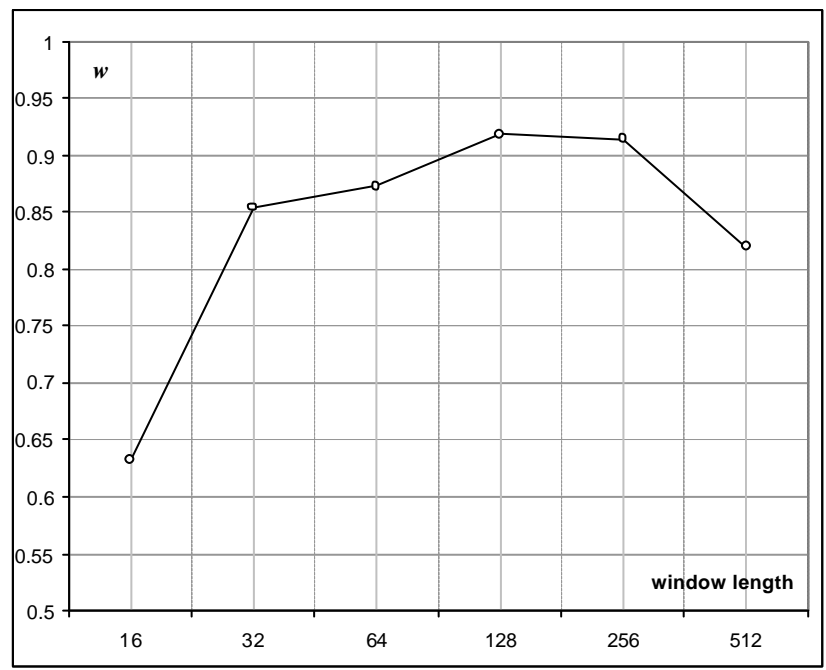

**Figure 3    Probability of correct recognition (spectral algorithm)**

It can be seen from the figure that probability value has a distinct maximum at FFT size equal to 128-256 samples, however, the sharp drop in the value of *w* implies that the optimal value has to be reestimated for each particular dictionary individually. The decreased probability of correct recognition for larger FFT sizes relates to the redundancy of spectral components in the segments of the recognized word. As soon as speech signals were found out to have only 4-5 significant formats [12], larger FFT sizes tend to mask significant formants with spurious spectral components leading to poor recognition.

Figure 4 shows how the probability of recognition of digit "4" (filter algorithm) relates to the order of AR model.

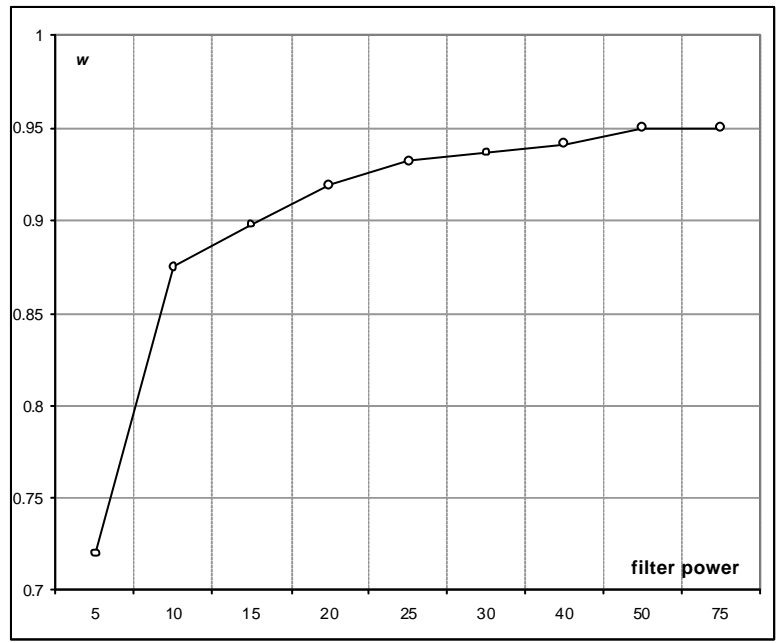

**Figure 4**     **Probability of correct recognition (filter algorithm)**

It can be seen that the probability is monotone increasing with respect to AR order and tends to stabilize for orders over 40. The shapes of the graphs suggest that the optimal order of AR model given the increasing computational requirements is approximately 30-50. A relatively low probability of correct filter recognition compared to other algorithms arises from the specifics of autoregressive spectral estimates that are characterized by sharper peaks with high amplitude. Such estimates are generally biased and provide poor approximation (e.g. compared to FFT estimates) for minor speech formants therefore degrading the reliability of recognition. The undoubted advantage of the algorithm, however, is the extremely low sensitivity to the choice of AR order.

Another supporting evidence of LID algorithms efficiency in speech recognition is high recognition quality in the case of a different speaker [14-16].

In order to give comparative grounds for the discussed results the speech signal database was used to test other existing and effective recognition algorithms. The comparative data is presented in table 1 listing the averaged probabilities of correct recognition for all three methods compared to autoregressive cepstral algorithm and Mel Scale Filter Bank (MSFB). Cepstral algorithm is based on AR estimates of cepstrum [17] and the generalized exponential distance metric [17] (Hermansky-Junqua). MSFB is based on a nonlinear frequency scale bank of filters [18], a set of cepstral coefficients calculated by Discrete Cosine Transform [19] and the Euclid's distance metric.

The table is presented on a conventional scale with respect to the value of the optimization parameter.

**Table 1.**

| Relative parameter value | Correlation algorithm | Spectral algorithm | Filter algorithm | Cepstral algorithm | MSFB |
|---|---|---|---|---|---|
| 1 | 0,90 | 0.73 | 0.85 | 0.66 | 0,72 |
| 2 | 0,93 | 0.84 | 0.87 | 0.72 | 0,76 |
| 3 | 0,96 | 0.88 | 0.88 | 0.75 | 0,80 |
| 4 | 0,98 | 0.93 | 0.92 | 0.81 | 0,85 |
| 5 | 0,99 | 0.91 | 0.95 | 0.83 | 0,88 |
| 6 | 0,98 | 0.85 | 0.95 | 0.86 | 0,88 |

It can be seen from the table that the suggested LID algorithms provide better recognition over the optimal range of optimization parameter values.

## Summary

The discussed LID approaches to speech recognition allowed to suggest three respective LID algorithms that were investigated and tested in a task of experimental recognition. Final results characterize both computational requirements of the algorithms and their efficiency in recognition. The article provides the results of experimental speech modeling and the comparative analysis of the three algorithms, pointing out the advantages and the downsides of each; some recommendations on model parameters are also made.

Additional analysis of other existing methods (AR cepstral coefficients, MSFB) was carried out to provide comparative data of recognition efficiency. Further comparison revealed a better reliability of LID algorithms in the task of recognition for the experimental dictionary database.